\documentclass[10pt,twocolumn,letterpaper]{article}

\usepackage{cvpr}              % To produce the CAMERA-READY version
\usepackage[accsupp]{axessibility}
\definecolor{cvprblue}{rgb}{0.21,0.49,0.74}
\usepackage[pagebackref,breaklinks,colorlinks,allcolors=cvprblue]{hyperref}
\usepackage{xcolor}
\usepackage[table]{xcolor}

\definecolor{iccvblue}{rgb}{0.21,0.49,0.74}

\def\paperID{} % *** Enter the Paper ID here
\def\confName{CVPR}
\def\confYear{2026}

\title{LiveGesture: Streamable Co-Speech Gesture Generation Model
}

\author{
Muhammad Usama Saleem\textsuperscript{1} \quad
Mayur Jagdishbhai Patel\textsuperscript{1} \quad
Ekkasit Pinyoanuntapong\textsuperscript{1} \quad
Zhongxing Qin\textsuperscript{1} \quad \\
Li Yang\textsuperscript{1} \quad
Hongfei Xue\textsuperscript{1} \quad
Ahmed Helmy\textsuperscript{1} \quad
Chen Chen\textsuperscript{2} \quad
Pu Wang\textsuperscript{1} \\
\textsuperscript{1}University of North Carolina,
\textsuperscript{2}University of Central Florida \\
\texttt{msaleem2@charlotte.edu}
}

\newcommand{\partitle}[1]{\smallskip \noindent \textbf{#1}.}

\usepackage{pifont,xcolor}
\newcommand{\cmark}{\textcolor{green!60!black}{\ding{51}}}
\newcommand{\xmark}{\textcolor{red!70!black}{\ding{55}}}

\begin{document}

\makeatletter
\let\@oldmaketitle\@maketitle % Store \@maketitle
\renewcommand{\@maketitle}{
  \@oldmaketitle % Insert the original title
  \vspace{1em} % Space between title and image
  \centering
  \includegraphics[width=0.9\linewidth]{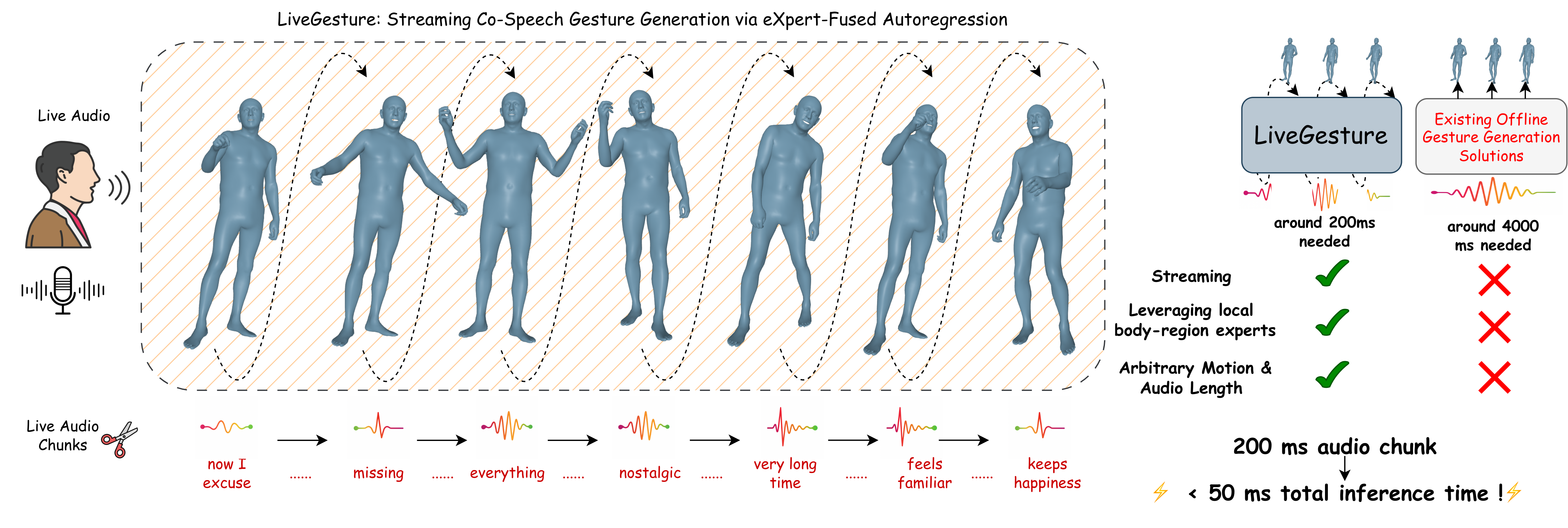}
 \captionof{figure}{
    LiveGesture overview. Given live audio chunks, our framework generates full-body SMPL-X motion online with zero look-ahead. 
    A streamable SVQ motion tokenizer and a hierarchical eXpert-fused autoregressive model (region-wise AR eXperts plus causal spatial--temporal fusion) 
    enable low-latency ($<\!50$ ms per 200 ms chunk) generation of diverse, beat-synchronous gestures over arbitrary-length speech, 
    in contrast to prior offline gesture methods that require full utterances and incur much higher latency.%
  }

  \label{fig:landing}

  \bigskip
}
\makeatother

\maketitle
\begin{abstract}

We propose LiveGesture, the first fully streamable, speech-driven full-body gesture generation framework that operates with zero look-ahead and supports arbitrary sequence length. Unlike existing co-speech gesture methods—which are designed for offline generation and either treat body regions independently or entangle all joints within a single model—LiveGesture is built from the ground up for causal, region-coordinated motion generation. \emph{LiveGesture} consists of two main modules: the Streamable Vector-Quantized Motion Tokenizer (SVQ) and the Hierarchical Autoregressive Transformer (HAR). The SVQ tokenizer converts the motion sequence of each body region into causal, discrete motion tokens, enabling real-time, streamable token decoding. On top of SVQ, HAR employs region-eXpert autoregressive (xAR) transformers to model expressive, fine-grained motion dynamics for each body region. A causal spatio-temporal fusion module (xAR-Fusion) then captures and integrates correlated motion dynamics across regions. Both xAR and xAR-Fusion are conditioned on live, continuously arriving audio signals encoded by a streamable causal audio encoder. To enhance robustness under streaming noise and prediction errors, we introduce autoregressive masking training, which leverages uncertainty-guided token masking and random region masking to expose the model to imperfect, partially erroneous histories during training. Experiments on the BEAT2 dataset demonstrate that LiveGesture produces coherent, diverse, and beat-synchronous full-body  gestures in real time, matching or surpassing state-of-the-art offline methods under true zero–look-ahead conditions. Project website: \url{https://m-usamasaleem.github.io/publication/LiveGesture/LiveGesture.html}.

\end{abstract}    
%\vspace{-15}
\section{Introduction}
\label{sec:intro}

Gestures naturally emerge alongside speech, emphasizing key ideas, conveying intent, and grounding abstract concepts in physical space. Far from being cosmetic, these non-verbal behaviors are central to how humans communicate and strongly influence perceived engagement and conversational naturalness. As virtual humans and embodied AI agents become more common in assistants, VR/AR avatars, telepresence, and content creation~\cite{tang2025generative,adaptive,huang2024modelingdrivinghumanbody,song2024texttoon,song2024tri,song2021talking}, generating believable, real-time co-speech gestures is becoming a core requirement.

Recent co-speech gesture models~\cite{mambatalk,liu2025contextualgesturecospeechgesture,liu2023emage,yi2022generating} synthesize full-body motion using either continuous trajectories~\cite{habibie2021learning} or discrete motion tokens~\cite{liu2023emage,yi2022generating}. While they produce locally plausible movements, their decoupled region representations often fail to capture fine-grained interdependencies across body parts, leading to gestures that lack holistic full-body coordination. Most systems rely on \emph{offline} diffusion-based~\cite{diffsheg,chen2024syntalker} or autoregressive~\cite{yi2022generating,liu2023emage} architectures that assume access to complete utterances or long audio/text contexts, incurring high latency and preventing truly interactive use. GestureLSM~\cite{liu2025gesturelsm} moves toward fast inference with a lightweight architecture, but remains \emph{non-streamable}, requiring full speech segments and preventing incremental updates as audio arrives.

In this work, we introduce \emph{LiveGesture}, to our knowledge the first \emph{fully streamable}, zero–look-ahead co-speech full-body gesture generation framework. LiveGesture comprises two core components: the Streamable Vector-Quantized Motion Tokenizer (SVQ) and the Hierarchical Autoregressive Transformer (HAR). The SVQ tokenizer encodes each body region’s motion sequence into causal, discrete motion tokens, enabling real-time, low-latency decoding. Built on top of SVQ, HAR employs region-eXpert autoregressive (xAR) transformers to model expressive and fine-grained motion dynamics for individual regions. A causal spatiotemporal fusion module (xAR-Fusion) further integrates inter-region dependencies, capturing coherent, whole-body coordination. Both xAR and xAR-Fusion operate under strict causality, conditioned on continuously arriving audio features extracted by a streamable causal audio encoder. To improve robustness against streaming noise and prediction drift, we introduce autoregressive masking training, which applies uncertainty-guided token masking and random region masking to expose the model to imperfect and partially corrupted histories. Evaluated on the BEAT dataset, \emph{LiveGesture} generates coherent, diverse, and beat-synchronous full-body SMPL-X gestures in real time—achieving state-of-the-art performance under true zero–look-ahead conditions.
Our contributions are summarized as follows:
% \begin{itemize}
%   \item We propose LiveGesture, the first fully streamable co-speech full-body gesture generation framework operating directly on SMPL-X motion.
%   \item We introduce a Streamable Asymmetric Motion Tokenizer (SVQ) and a hierarchical region-eXpert autoregressive transformer with PILOR adapters and causal spatio-temporal fusion (xAR-Fuse), enabling strictly causal, low-latency full-body gesture prediction from audio (and optional text).
%   \item We design a streaming-robust training strategy with hybrid uncertainty-guided and region masking to mitigate exposure bias, and show that LiveGesture  surpasses state-of-the-art offline methods while enabling real-time streaming in interactive avatar applications.
% \end{itemize}

\begin{itemize}
\item  We propose \emph{LiveGesture}, the first zero–look-ahead, fully streamable speech-driven full-body gesture generation framework supporting arbitrary sequence lengths, designed for real-time human–AI interaction.

\item We introduce a streamable motion tokenizer that converts continuous SMPL-X regional motion sequences into discrete latent motion tokens through an asymmetric architecture of bidirectional encoding, causal decoding, and token quantization, trained via a two-stage pretraining and quantization strategy to ensure strict causality, low latency, and stable representation learning. 

\item  We design a Hierarchical Autoregressive Transformer that consists of region-eXpert autoregressive (xAR) transformers that model expressive, fine-grained motion dynamics for each body region, a causal spatiotemporal fusion module (xAR-Fusion) that captures and integrates correlated motion dynamics across regions, and a streamable causal audio encoder that processes continuously arriving audio signals for live gesture generation.

\item Experiments on the \textbf{BEAT} dataset demonstrate that \emph{LiveGesture} produces coherent, diverse, and beat-synchronous SMPL-X gestures in real time, matching or surpassing state-of-the-art offline methods under true zero–look-ahead conditions.
\end{itemize}

\vspace{-5pt} 

\section{Related Works}
\vspace{-5pt} 
\partitle{Autoregressive Motion Generation}
Autoregressive models have shown remarkable success in language and image generation, as demonstrated by GPT \cite{GPT}, DALL-E \cite{DALL-E}, and VQ-GAN \cite{vqgan, Hierarchical-vqvae, vit-vqgan11}. Following this, Masked Motion Models \citep{MMM, momask} have adopted discrete motion token sequences from pretrained codebooks in human motion generation by decoding them in parallel. However, these models require generating all sequences at once and require a predefined motion length. 
In contrast, methods such as T2M-GPT \cite{T2M-GPT}, AttT2M \cite{AttT2M}, and BAMM \cite{BAMM} also rely on codebook tokens but train their models autoregressively to predict motion token sequences token by token. Similarly, MotionGPT \cite{MotionGPT} generates motion tokens autoregressively, but integrates both language and motion tokens into a unified encoder, enabling a broader range of motion-related tasks. However, these approaches cannot achieve streamable generation because their motion-token decoders rely on a bidirectional decoder, future tokens influence the decoding of current ones.

% While, DartControl \cite{Zhao:DartControl:2025} generates motion in chunked rollouts conditioned on past frames rather than streaming the output token by token.

% DartControl \cite{Zhao:DartControl:2025} further employs an autoregressive rollout to generate motion sequences from real-time text inputs

% Most existing approaches to co-speech gesture generation rely on skeleton- or joint-level pose representations. 
\partitle{Co-speech Gesture Generation}
Early work by \cite{ginosar2019gestures} employs an adversarial model to predict hand and arm poses directly from audio and uses conditional video generation techniques based on pix2pixHD~\cite{wang2018pix2pixHD} and EverybodyDanceNow~\cite{EverybodyDanceNow}. More recent studies~\cite{liu2022learning,Deichler_2023,xu2023chaingenerationmultimodalgesture, liu2024tango, zhang2024kinmokinematicawarehumanmotion, liu2025contextualgesturecospeechgesture,song2021fsft,liu2025intentionalgesturedeliverintentions,liu2025semgessemanticsawarecospeechgesture} explore generative models to derive audio-conditioned gesture generations. For example, HA2G~\cite{liu2022learning} constructs both high-level and low-level audio–motion embeddings to guide gesture synthesis. TalkShow~\cite{yi2022generating} models joint body and hand motion dynamics for talk-show videos. CaMN~\cite{liu2022beat} and EMAGE~\cite{liu2023emage} leverage GPT-like decoders for unified face and body modeling. Several methods focus on improving generation quality or efficiency. MambaTalk~\cite{mambatalk} accelerates gesture synthesis through the state space model architecture. DiffSHEG~\cite{diffsheg},  SynTalker~\cite{chen2024syntalker}, and GestureLSM \cite{liu2025gesturelsm} adopt diffusion-based pipelines for co-speech gesture synthesis. However, none of the aforementioned methods support live streamable gesture generation because they require access to the complete text and speech input before generating gestures.

\begin{figure}[ht]
    \centering
    \includegraphics[width=0.9\linewidth]{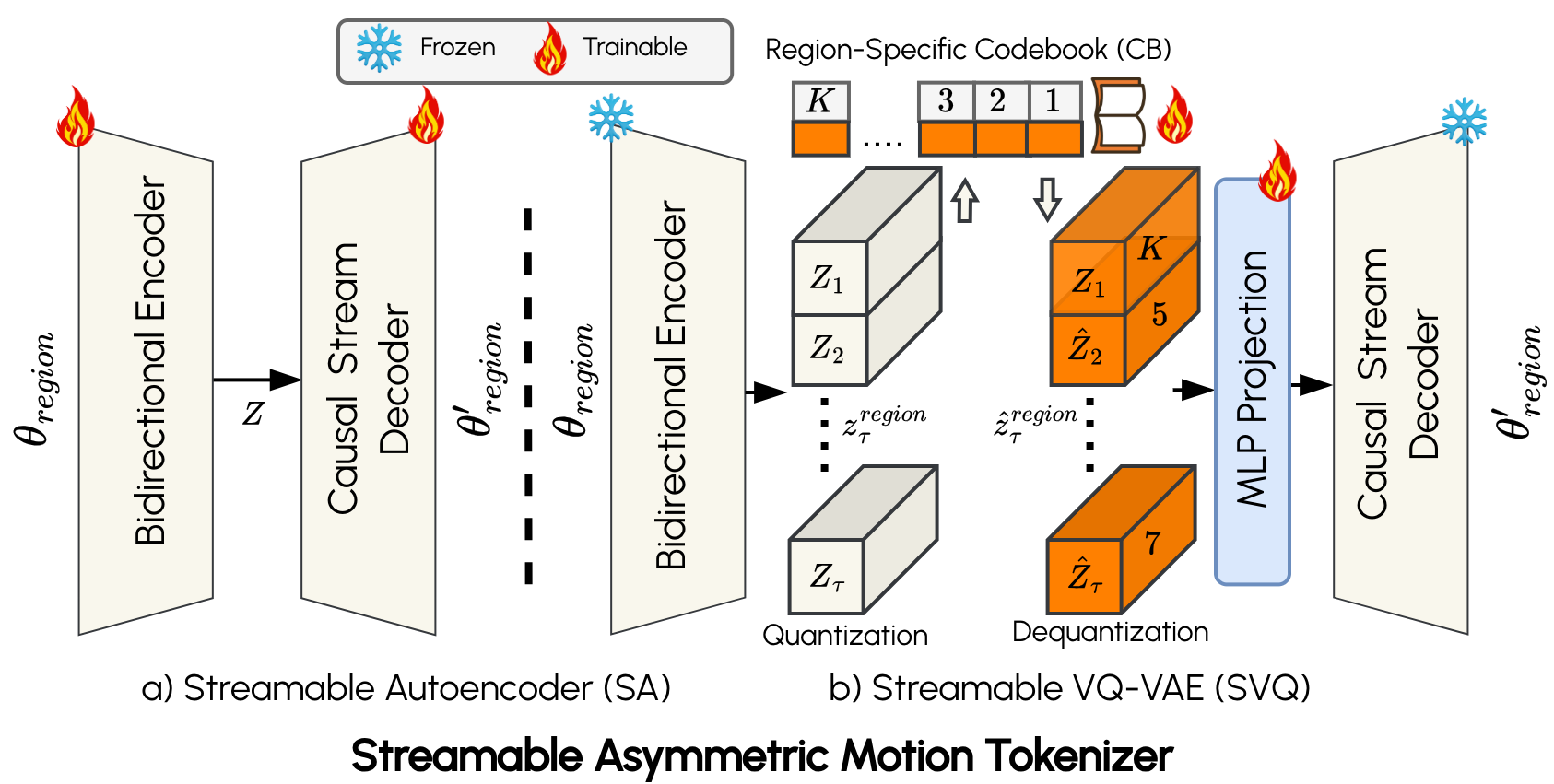}

\caption{ Overview of the Streamable Asymmetric Motion Tokenizer. (a) For each body region, a bidirectional encoder and causal stream decoder learn low-rate latent sequences. (b) Freezing this autoencoder, the Streamable VQ-VAE (SVQ) adds a region-specific codebook and projection head to quantize latents into discrete, time-synchronous motion tokens that remain compatible with strictly causal decoding for streaming. }

    \label{fig:stage1_vqvae}
\end{figure}

\begin{figure*}[ht]
    \centering
    \includegraphics[width=0.85\linewidth]{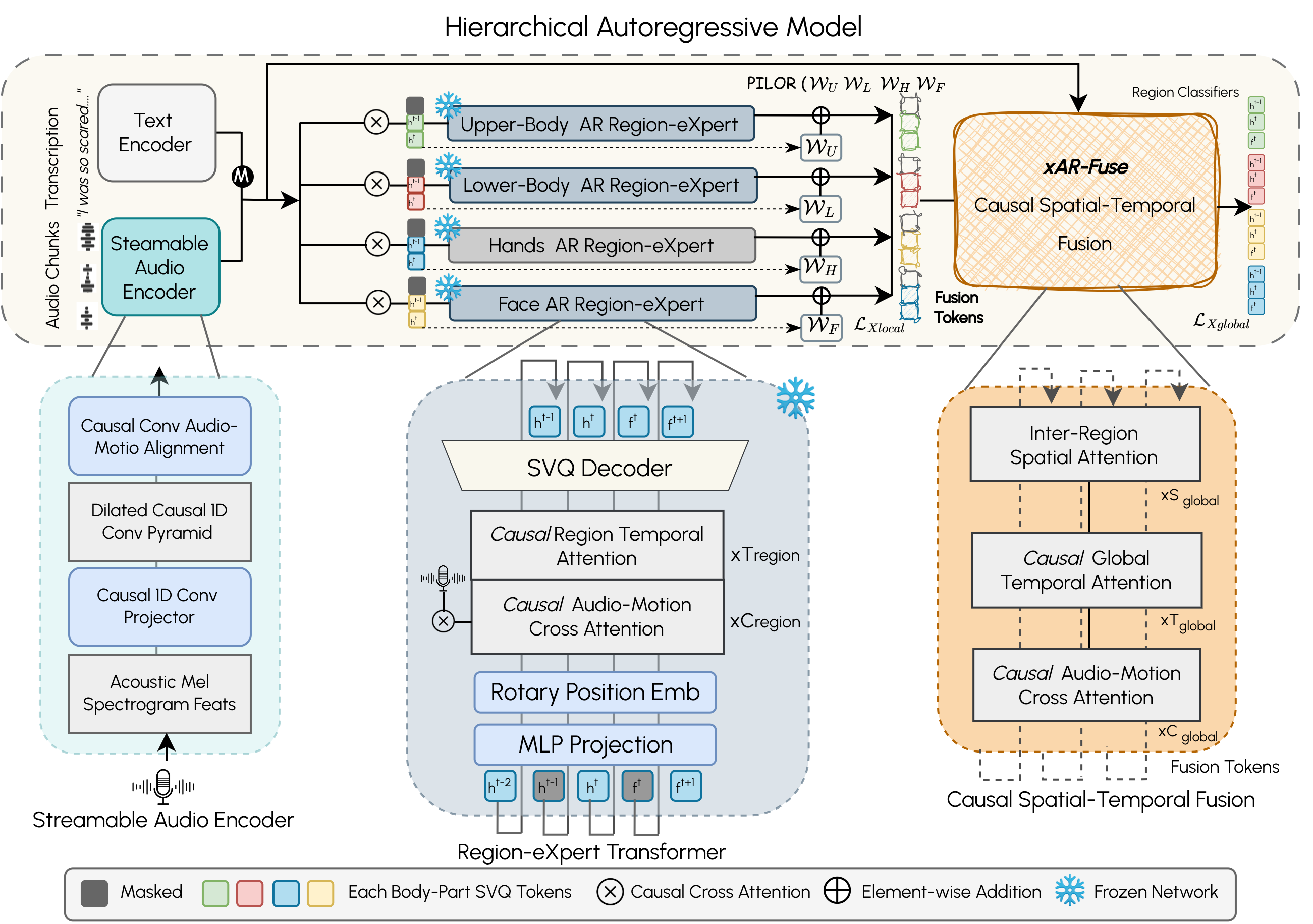}

\caption{
Overview of the Hierarchical Autoregressive Model in \emph{LiveGesture}. A streamable audio encoder and optional text encoder provide causal audio/text tokens to four local AR Region-eXperts (upper body, lower body, hands, face), each modeling its own SVQ motion token stream. Their frozen states are adapted by per-region Pre-Infusion Adapters (PILOR) and fused by xAR-Fuse, a causal spatial–temporal transformer with audio–motion cross-attention, global temporal attention, and inter-region spatial attention that predicts next-step SVQ tokens for zero–look-ahead streaming full-body gesture generation.}

    \label{fig:overview}
\end{figure*}
\vspace{-5pt}
\section{Proposed Method: LiveGesture} 
\vspace{-5pt}
\partitle{Problem Formulation}
We study \emph{streaming} co-speech full-body gesture generation with \emph{zero look-ahead}. At time \(t\), the model receives only the recent motion history \(\mathbf{S}_t = [\mathbf{q}_{t-H+1},\ldots,\mathbf{q}_{t-1}]\), the current audio token \(a_t\), and optional online transcript tokens \(w_t\), and must predict the next full-body pose gesture \(\mathbf{q}_t\). Formally, the gesture generation \(f_\Theta\) is strictly causal, $\hat{\mathbf{q}}_t = f_\Theta(\mathbf{S}_t, a_t, w_t),$. Unlike offline formulations that access future audio or motion, this setup enforces real-time multimodal conditioning and requires region-coordinated, temporally smooth motion aligned with speech rhythm and content under strict latency constraints.

\partitle{Method Overview}
As illustrated in Figure~\ref{fig:stage1_vqvae} and Figure~\ref{fig:overview}, \emph{LiveGesture} comprises two main modules, \emph{Streamable Vector-Quantized Motion Tokenizer (SVQ)} and the \emph{Hierarchical Autoregressive Transformer (HAR)}, for strictly causal, zero–look-ahead inference. First, the motion tokenizer converts the motion sequence of each body region into causal and discrete motion tokens to support real-time streamable token decoding. On top of SVQ, HAR first employs region-eXpert autoregressive (xAR) transformers to
learn expressive and fine-grained motion dynamics of each 
body region. Then, the causal spatial-temporal fusion (xAR-fusion) model captures and fuses correlated motion dynamics among different body regions. Both xAR and xAR-fusion models are conditioned on the live-continuously arriving audio signals encoded by a streamable audio encoder.

\subsection{Streamable Vector-Quantized Motion Tokenizer}
\vspace{-5pt} 
\label{sec:svq}

We propose a per-region streamable motion tokenizer that converts the continuous motion sequence of each SMPL-X body region (lower body, upper body, hands, and head)~\cite{pavlakos2019expressive} into discrete latent motion tokens while preserving strict causality and low latency. The tokenizer adopts an asymmetric architecture comprising three key components: (1) a bidirectional encoder that projects raw motion sequence into a latent space capturing rich bidirectional dependencies; (2) a causal decoder that reconstructs motion in a strictly streamable and causal manner from the latent representation; and (3) a token quantizer that discretizes continuous motion embeddings into compact latent tokens, enabling autoregressive motion generation through cross-entropy training. Previous research shows that simultaneous training of encoder and quantizer can cause token and embedding collapses, leading to degradation in the quality of learned representations. To mitigate this problem, we adopt a two-stage training strategy: Asymmetric Autoencoder Pretraining and Quantization Learning. 
 
\partitle{Asymmetric Autoencoder Pretraining}
In the first stage, we train separate \emph{Streamable Autoencoders} (SA) for each region's SMPL-X parameters. Let \(\boldsymbol{\theta}^{\text{SMPL-X}}_t \in \mathbb{R}^{D}\) denote the full SMPL-X vector at original motion frame time \(t \in \{1,\dots,T_{\text{f}}\}\), partitioned into region-specific subsets \(\boldsymbol{\theta}^{\text{region}}_t \in \mathbb{R}^{d_{\text{region}}}\). For each region, the input sequence \(\{\boldsymbol{\theta}^{\text{region}}_t\}_{t=1}^{T_{\text{f}}}\) is encoded by a 1D convolutional encoder \(E\) into a downsampled latent sequence
\[
z^{\text{region}} = \{ z_\tau \}_{\tau=1}^{T} = E(\{\boldsymbol{\theta}^{\text{region}}_t\}_{t=1}^{T_{\text{f}}}),
\quad T = T_{\text{f}}/4,
\]
using strided convolutions to reduce the frame rate by a factor of 4 and obtain a compact representation for fast streaming inference. The encoder is bidirectional, aggregating past and future context, while the corresponding \emph{Causal Stream Decoder} \(D_{\text{CS}}\) is strictly causal and reconstructs \(\hat{\boldsymbol{\theta}}^{\text{region}}_t\) from the latent sequence \(z^{\text{region}}\). This asymmetric design yields a globally coherent latent space while enforcing the causal decoding required for streaming. Each SA is trained with a reconstruction loss
\[
\mathcal{L}_{\text{AE}} = \lambda_{\text{AE}} \mathcal{L}_{\text{recon}}.
\]
\vspace{-5pt} 
\partitle{Quantization Learning}
In the second stage, we convert the continuous latents \(z^{\text{region}} = \{ z_\tau \}_{\tau=1}^{T}\) into discrete tokens, while keeping the encoder \(E\) and decoder \(D_{\text{CS}}\) frozen so that the temporal geometry and streamability learned in Stage~1 are preserved. For each region, we introduce a region-specific codebook
\[
C^{\text{region}} = \{ c_k \}_{k=1}^{K},
\]
and vector-quantize \(z^{\text{region}}\) via nearest-neighbor assignment, obtaining dequantized embeddings
\[
\hat{z}^{\text{region}} = \{ \hat{z}_\tau \}_{\tau=1}^{T},
\]
whose indices define time-synchronous motion tokens at the downsampled rate. A lightweight projection head \(W^{\text{region}}\), implemented as a small MLP, is applied after quantization to map each \(\hat{z}_\tau\) into the latent space expected by the frozen decoder, yielding $\tilde{z}^{\text{region}} = \{ \tilde{z}_\tau \}_{\tau=1}^{T},
\quad
\tilde{z}_\tau = W^{\text{region}}(\hat{z}_\tau)$

This projection serves as a learned adapter between the discrete codebook space and the original autoencoder latent space: it allows the codebook to remain compact and region-specific, while the decoder operates in a richer, globally learned space, and it absorbs quantization artifacts so that codebook updates do not disrupt the temporal structure encoded in \(D_{\text{CS}}\). The projected sequence \(\tilde{z}^{\text{region}}\) is then decoded causally to reconstructed SMPL-X parameters, $\hat{\boldsymbol{\theta}}^{\text{region}} = D_{\text{CS}}(\tilde{z}^{\text{region}}).$
% \[
% \hat{\boldsymbol{\theta}}^{\text{region}} = D_{\text{CS}}(\tilde{z}^{\text{region}}).
% \]
During this stage, only the codebook \(C^{\text{region}}\) and projection head \(W^{\text{region}}\) are updated. We use an L1 reconstruction term on SMPL-X region parameters and a standard VQ codebook loss \(\mathcal{L}_{\text{cb}}(z^{\text{region}}, e^{\text{region}})\)~\cite{van2017neural}, where \(e^{\text{region}}\) denotes the selected codebook embeddings, leading to
\[
\begin{aligned}
\mathcal{L}_{\text{stage2}} =
&\ \lambda_{\text{rec}} \Bigl\lVert \boldsymbol{\theta}^{\text{region}}
- D_{\text{CS}}(W^{\text{region}}(\hat{z}^{\text{region}})) \Bigr\rVert_1 \\
&\ + \lambda_{\text{cb}} \,\mathcal{L}_{\text{cb}}(z^{\text{region}}, e^{\text{region}}).
\end{aligned}
\]
EMA-style codebook updates and occasional resets~\cite{esser2021taming} keep the codebook well conditioned, and the resulting SVQ tokens are discrete, time-synchronous, region-wise representations that remain compatible with the strictly causal decoder and are well suited for downstream streaming autoregressive SMPL-X gesture generation.
% \vspace{-5pt} 
\subsection{Hierarchical Autoregressive Model}
The hierarchical autoregressive model consists of  (1) region-eXperts, which are temporal causal transformers that learn expressive and fine-grained motion dynamics of each body region, (2) a global fusion model, which is a spatial-temporal causal transformer that captures the coherent and dependent dynamics among different body regions, and (3) a streamable audio encoder, which is a causal convolution network that enables live gesture generation conditioned on continuously arriving audio signals. 
\subsubsection{Region-eXpert Autoregressive Transformer (xAR)}
Local region experts learn audio-driven motion dynamics for individual body parts, as different regions exhibit distinct motion distributions. For example, the upper body often produces large-scale movements, while the hands perform fine-grained, high-frequency gestures. In particular, we divide full-body into four regions $
\mathcal{R} = \{\text{upper body}, \text{lower body}, \text{hands}, \text{face}\}$. For each region \(r \in \mathcal{R}\), we maintain a stream of motion tokens \(\{x^{r}_t\}_{t=1}^{T}\) and aligned audio tokens \(\{a_t\}_{t=1}^{T}\), with optional text tokens \(\{w_t\}_{t=1}^{T}\) when available. In this stage, each eXpert is trained independently, but all share the same audio encoder so they learn to respond to a common audio representation. At time step \(t\), the eXpert for region \(r\) receives a causal window of past region tokens \(\{x^{r}_{t-h},\ldots,x^{r}_{t-1}\}\) and audio tokens \(\{a_{t-h},\ldots,a_{t}\}\). Motion tokens are embedded by an MLP projection and augmented with rotary positional embeddings, yielding a continuous sequence that is stable under streaming shifts. This sequence is processed by a small stack of causal Transformer blocks that interleave (i) causal audio–motion cross attention, where region tokens attend only to past and current audio (and text) tokens, which enables rhythm-and-gesture alignment, and (ii) causal temporal self-attention, which captures region-specific inter-token correlations. The output motion embedding \(h^{r}_t\) is fed into a token classifier that models audio-condition categorical motion distribution. 

% This factorized design lets each region specialize in its own articulation statistics without being constrained by full-body modeling, and keeps per-region computation small enough for low-latency streaming. Using a shared LASER encoder across regions ensures that all eXperts are aligned in the same audio feature space, which is critical for the subsequent fusion stage.
\vspace{-5pt} 
\subsubsection{Causal Spatial-Temporal Fusion (xAR-Fuse)}

Local eXperts learn rich region-specific motion distributions but do not explicitly enforce full-body coordination. The causal fusion transformer operates on top of the frozen eXperts to model inter-region spatiotemporal correlations while preserving strict causality. At each time step \(t\), we collect the hidden embeddings \(\{h^{r}_t\}_{r \in \mathcal{R}}\) from all region eXperts. Since these embeddings are generated by independently trained networks, they are not naturally aligned for joint fusion. To bridge this gap, we introduce a lightweight residual adapter for each region, i.e., $\Delta h^{r}_t = \mathcal{W}_r h^{r}_t, \quad
\tilde{h}^{r}_t = h^{r}_t + \Delta h^{r}_t.$
% \[
% \Delta h^{r}_t = \mathcal{W}_r h^{r}_t, \quad
% \tilde{h}^{r}_t = h^{r}_t + \Delta h^{r}_t.
% \]
This adapter gently aligns the output embeddings from different eXperts into a shared fusion space with minimal parameters and computational overhead. The adapted embeddings \(\{\tilde{h}^{r}_t\}_{r \in \mathcal{R}}\) are passed through the fusion transformer, implemented as a stack of causal Transformer blocks. Each block is factorized into three attention layers that separately address audio-motion coupling, temporal refinement, and spatial coordination. A causal audio–motion cross-attention layer allows each region to query the current audio (and text) tokens, reinforcing beat and semantic alignment at every step. The inter-region spatial-attention layer, coupled with the causal global temporal-attention layer, explicitly and efficiently captures whole-body spatiotemporal coordination such as arm–torso coupling or mirrored hand gestures. Instead of sharing a single MLP token classifier, the output embeddings from the fusion transformer are finally fed into region-specific token classifiers to enhance fine-grained regional expressiveness.

\subsubsection{Streamable Audio Encoder}
To enable live gesture generation conditioned on continuously arriving audio signals, we design a causal audio encoder. First, an acoustic mel-spectrogram feature extraction module converts raw waveforms into log-mel frames using a short analysis window and small hop, chosen to match the latency budget of downstream streaming.  These features are processed by a causal 1D convolutional projection module with left-only padding that enforces causality from the first layer. A dilated causal convolutional pyramid then captures multi-scale temporal structures in rhythm and prosody, expanding the receptive field without increasing latency. Finally, a causal audio–motion alignment module aggregates pyramid activations within each motion step using temporal striding, producing audio tokens \(a_t\)  at the same frame rate as motion tokens. Each audio token depends only on past and current acoustic evidence, yielding multi-scale, low-latency, and temporally aligned representations ideal for zero–look-ahead streaming gesture generation.

\vspace{-5pt} 
\subsection{Autoregressive Masked Training}

Our model is trained in two stages: first learn audio-driven autoregressive motion dynamics for each region, and then train the fusion transformer with a hybrid masking strategy. 

\partitle{Stage 1: Local autoregressive modeling.}
For each region \(r \in \mathcal{R}\), the local eXpert defines a causal token-level model
\[
p_{\phi}^{r}(x^{r}_{1:T} \mid a_{1:T}, w_{1:T})
= \prod_{t=1}^{T} p_{\phi}^{r}\bigl(x^{r}_t \mid x^{r}_{1:t-1}, a_{1:t}, w_{1:t}\bigr),
\]
where \(x^{r}_t\) is the motion token for region \(r\) at time \(t\), \(a_t\) is the audio token, and \(w_t\) is the optional text token. During training, we teacher-force the ground-truth history and optimize a standard autoregressive cross-entropy loss
\[
\mathcal{L}_{\text{local}}
= - \sum_{r \in \mathcal{R}} \sum_{t=1}^{T}
\log p_{\phi}^{r}\bigl(x^{r}_t \mid x^{r}_{1:t-1}, a_{1:t}, w_{1:t}\bigr),
\]
optionally combined with a small reconstruction loss in pose space. To mitigate \emph{exposure bias}, i.e., the mismatch between teacher-forced training histories and self-generated histories at inference time, we also inject small Gaussian noise into the embedded history tokens with a small probability \(p_{\text{noise}}\), so that local eXperts learn to cope with slightly corrupted context rather than relying on perfectly clean ground-truth sequences. This stage yields frozen local eXperts that provide well-calibrated next-token distributions and hidden states \(h^{r}_t\) for each region.

\partitle{Stage 2: Hybrid Masking for Robust Fusion}
To capture the inherent motion dynamics dependency among different body regions, a certain number of motion tokens from different regions at different time instances are masked out. The fusion transformer is trained to predict the next token based on partially corrupted motion sequence,  audio embeddings, and optional text transcript tokens.  The reconstruction probability of each motion token under motion corruptions is   
\[
p_{\theta}\bigl(x^{r}_t \mid \tilde{x}^{1:|\mathcal{R}|}_{1:t}, a_{1:t}, \tilde{w}_{1:t}\bigr),
\quad r \in \mathcal{M}_t,
\]
where \(\tilde{x}^{1:|\mathcal{R}|}_{1:t}\) represent corrupted motion sequence up to time \(t\) and \(\tilde{w}_{1:t}\) are the (possibly masked) text tokens. The training objective is to minimize the negative log-likelihood of the next motion token prediction
\[
\mathcal{L}_{\text{fuse}}
= - \sum_{t=1}^{T} \sum_{r \in \mathcal{M}_t}
\log p_{\theta}\bigl(x^{r}_t \mid \tilde{x}^{1:|\mathcal{R}|}_{1:t}, a_{1:t}, \tilde{w}_{1:t}\bigr).
\]

We adopt two types of masking strategies: uncertainty-guided token masking (UGM) and random region masking (RM). Together, they expose the fusion model to the imperfect, partially erroneous histories it will see at inference time, further reducing exposure bias. For token masking, the masking ratio (i.e., the percentage of tokens is masked) is controlled by a cosine scheduler over training steps. Concretely, let \(\lambda_{\text{UGR}}(s) \in [0,1]\) be a cosine-annealed masking ratio at training step \(s\). The $M_{\text{eff}}(s) = \bigl\lfloor \lambda_{\text{UGR}}(s)\, M_{\max} \bigr\rfloor$ least confident tokens based on their prediction probabilities are masked out.  Early in training, \(\lambda_{\text{UGR}}(s)\approx 0\), so masking ratio is low and the fusion transformer learns to inter-eXpert dependency under mostly clean inputs. As training progresses, \(\lambda_{\text{UGR}}(s)\) increases toward 1, and the model is gradually exposed to more aggressively corrupted motion sequences. In addition to token-level masking, we apply a \emph{region masking} scheme: with a small probability \(p_{\text{drop}}\) per training sequence, we select a region \(r_{\text{drop}} \in \mathcal{R}\) and treat its entire token trajectory \(\{x^{r_{\text{drop}}}_t\}_{t=1}^{T}\) as masked, forcing the fusion transformer to reconstruct the masked body region purely from audio and the remaining unmasked regions

% \vspace{-5pt} 
\partitle{Classifier-Free Logits Guidance}
To further enhance audio/text–motion alignment, we apply classifier-free guidance to the discrete logits of the token classifiers. During training, audio, text, or both modalities are randomly dropped with a small probability to enable the model to learn an unconditional prior.  At inference, the conditional and unconditional logits for each region \(r \in \mathcal{R}\) are combined as $\ell^{r}_{\text{guided}} =
\ell^{r}_{\text{uncond}} + \gamma\bigl(\ell^{r}_{\text{cond}} - \ell^{r}_{\text{uncond}}\bigr),$
where \(\gamma \geq 1\) denotes the guidance scale.  The resulting guided logits are then passed through a softmax function to produce the token distribution used for sampling during inference.

.

\begin{figure}[ht]
    \centering
    \includegraphics[width=1\linewidth]{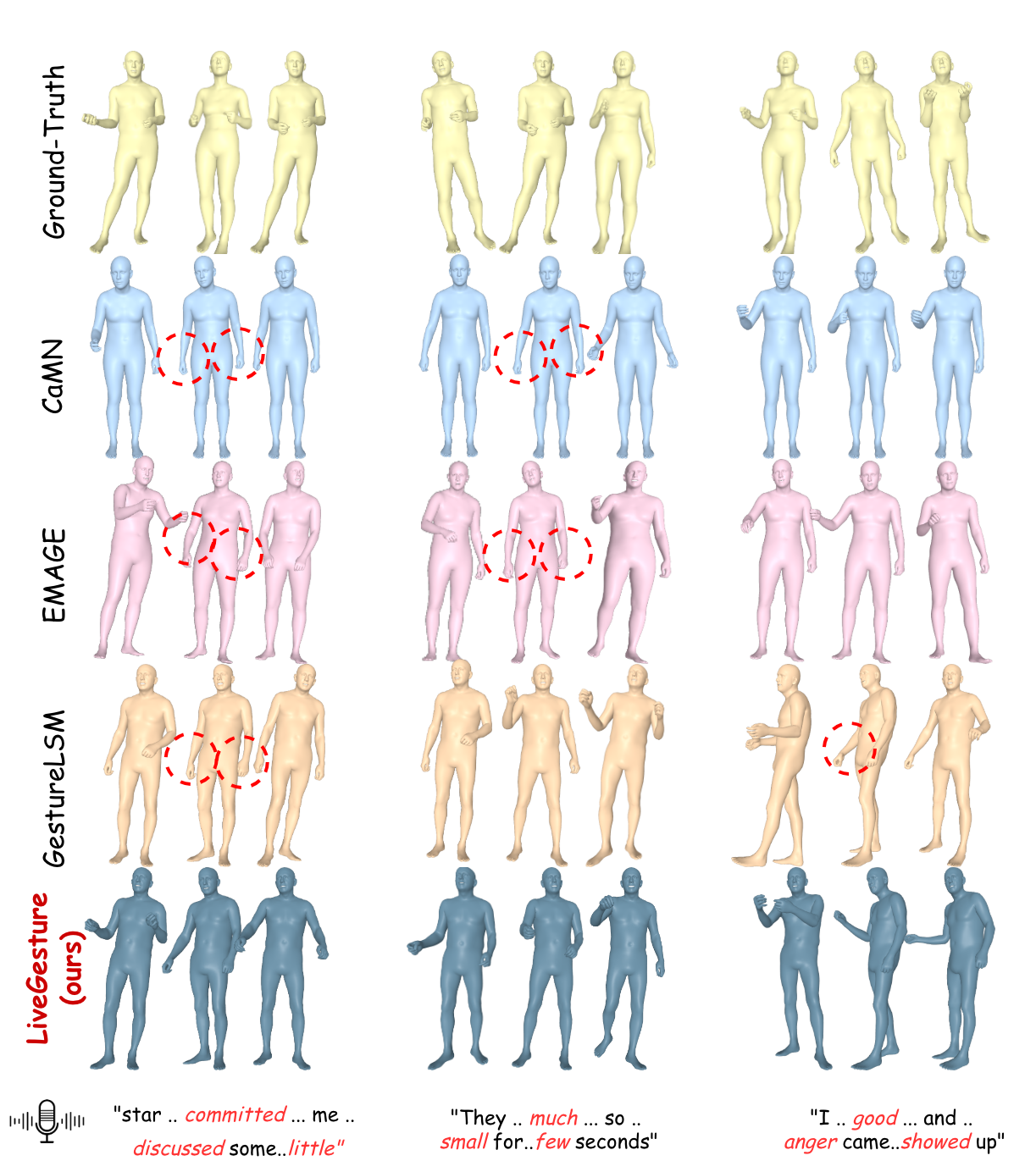}
    \caption{
        Qualitative comparison with state-of-the-art methods on BEAT. 
        LiveGesture generates more diverse, expressive full-body gestures that better follow speech rhythm, 
        while red circles mark representative failure cases of prior methods.
    }
    \label{fig:sota_comp}
\end{figure}

\vspace{-5pt} 
\section{Experiments}
\vspace{-5pt} 
\label{sec:experiment}

\partitle{Datasets} We train and evaluate LiveGesture on the BEAT2 corpus introduced by EMAGE~\cite{liu2023emage}, which contains 60 hours of SMPL-X full-body motion aligned with speech from 25 speakers (1,762 conversational clips, average length 65.66 s). The data cover diverse speaking styles and gesture behaviors. All experiments follow the official train/test split of EMAGE~\cite{liu2023emage} for fair comparison.

\partitle{Evaluation Metrics} We assess gesture quality in terms of realism, diversity, audio–motion alignment, and facial accuracy. Realism is measured by Fréchet Gesture Distance (FGD)~\cite{yoon2020speech}; diversity (Div.)~\cite{li2021audio2gestures} is the average L1 distance between generated clips for the same audio; audio–motion synchronization is evaluated by Beat Constancy (BC)~\cite{li2021ai}; and facial accuracy is measured by vertex MSE between predicted and ground-truth SMPL-X face vertices~\cite{xing2023codetalker}. For readability, FGD and BC are scaled by \(10^{-1}\) and MSE by \(10^{-7}\) in all tables.

% \partitle{Datasets} We train and evaluate LiveGesture on the BEAT2 corpus introduced by EMAGE~\cite{liu2023emage}. BEAT2 provides 60 hours of high-quality SMPL-X-based full-body motion aligned with speech from 25 speakers (12 female, 13 male), comprising 1,762 conversational sequences of average length 65.66 seconds. Each sequence records spontaneous responses to everyday questions, giving a wide range of speaking styles and gesture behaviors. To ensure comparability with prior work, all experiments follow the official train/validation/test split protocol of EMAGE~\cite{liu2023emage}.

% \partitle{Evaluation Metrics} We evaluate LiveGesture in terms of gesture realism, diversity, audio–motion alignment, and facial accuracy. Gesture realism is measured by Fréchet Gesture Distance (FGD)~\cite{yoon2020speech}, which compares the feature distributions of generated and real motions. Diversity (Div.)~\cite{li2021audio2gestures} is computed as the average L1 distance between pairs of generated gesture clips for the same audio, capturing variation beyond a single deterministic trajectory. Speech–gesture synchronization is assessed with Beat Constancy (BC)~\cite{li2021ai}, which measures how consistently motion beats align with prosodic accents. For facial motion, we report vertex Mean Squared Error (MSE)~\cite{xing2023codetalker} between predicted and ground-truth SMPL-X face vertices as a fine-grained accuracy metric. In all quantitative tables, FGD and BC are scaled by ($10^{-1}$) and MSE by ($10^{-7}$) for readability.

\begin{table*}[ht]
    \caption{
    State-of-the-art comparison on BEAT.  Best results are shown in \textbf{bold} and second best are \underline{underlined}. 
    The \textit{Streaming} column indicates whether the method supports zero-look-ahead streaming (\cmark) or is offline-only (\xmark); 
    Our method is the only streaming model while remaining superior in most of important metrics.}
    \label{tab:experiment}
    \centering
    \scalebox{0.85}{
    \begin{tabular}{l|lcccccc}
    \toprule
    Methods & Venue & Streaming & FGD ($\downarrow$) & BC ($\rightarrow$) & Diversity ($\uparrow$) & MSE ($\downarrow$) \\
    \midrule
    \rowcolor{gray!10}
    \multicolumn{7}{c}{\textbf{Offline Solutions}} \\
    \midrule
    HA2G~\cite{liu2022learning}                & CVPR'22    & \xmark & 12.32 & 0.677 &  8.626 &   --    \\
    DisCo~\cite{liu2022disco}                  & MM'22      & \xmark &  9.417 & 0.643 &  9.912 &   --    \\
    CaMN~\cite{liu2022beat}                    & ECCV'22    & \xmark &  6.644 & 0.676 & 10.86  &   --    \\
    TalkShow~\cite{yi2022generating}           & CVPR'23    & \xmark &  6.209 & 0.695 & 13.47  &  7.791 \\
    DiffSHEG~\cite{diffsheg}                   & CVPR'24    & \xmark &  7.141 & 0.743 &  8.21  &  9.571 \\
    ProbTalk~\cite{probtalk}                   & CVPR'24    & \xmark &  5.040 & 0.771 & 13.27  &  8.614 \\
    EMAGE~\cite{liu2023emage}                  & CVPR'24    & \xmark &  5.512 & 0.772 & 13.06  & 7.680 \\
    MambaTalk~\cite{mambatalk}                 & NeurIPS'24 & \xmark &  5.366 & \underline{0.781} & 13.05  & 7.680 \\
    SynTalker~\cite{chen2024syntalker}         & MM'24      & \xmark &  4.687 & 0.736 & 12.43  &   --    \\
    RAG-Gesture~\cite{mughal2025retrieving}        & CVPR'25 & \xmark  & 9.110 & 0.727 & 12.62  &   - \\
    RAG-Gesture (w/ Discourse) ~\cite{mughal2025retrieving}        & CVPR'25    & \xmark & 8.790 & 0.739 & 12.62 &   - \\

    GestureLSM~\cite{liu2025gesturelsm}        & ICCV'25    & \xmark & \textbf{4.247} & 0.729 & \underline{13.76}  &   \textbf{1.021} \\
    \midrule
    \rowcolor{gray!10}
    \multicolumn{7}{c}{\textbf{Streaming Solution (Ours)}} \\
    \midrule
    \rowcolor{green!8} \textbf{LiveGesture (ours)} 
        & Ours     
        & \cmark 
        & \underline{4.57} 
        & \textbf{0.794} 
        & \textbf{13.91} 
        & \underline{1.241} \\
    \bottomrule
    \end{tabular}}
    \vspace{-0.3cm}
\end{table*}
\vspace{-5pt} 

Table~\ref{tab:experiment} compares LiveGesture with recent state-of-the-art co-speech gesture models on BEAT2. Although it is the \emph{only} zero–look-ahead streaming method, LiveGesture remains competitive with or superior to offline systems: it achieves near-best Fréchet Gesture Distance (FGD; 4.57 vs.\ 4.25 for GestureLSM), the best Beat Constancy (BC = 0.794), the highest Diversity (13.91), and the second-lowest facial MSE (1.241 vs.\ 1.021), indicating tight rhythm alignment, rich motion variation, and high-fidelity full-body and facial motion under strict streaming constraints. We attribute this to three factors: (i) the streamable asymmetric motion tokenizer (SVQ), which learns a globally coherent latent space while decoding strictly causally, providing clean low-rate motion tokens; (ii) region-wise AR eXperts plus xAR-Fuse, which align frozen eXpert states and apply causal spatio-temporal attention for coherent whole-body coordination; and (iii) hybrid masked training on top of a strictly causal audio encoder, which exposes the model to realistic corrupted histories and yields robust audio–motion coupling. Together, these choices allow LiveGesture to achieve extremely low First-Token Latency ($250$ ms), the amount of time a model takes to generate the first motion frame in its response after receiving the first audio token.

\begin{figure}[ht]
    \centering
    \includegraphics[width=0.7\linewidth]{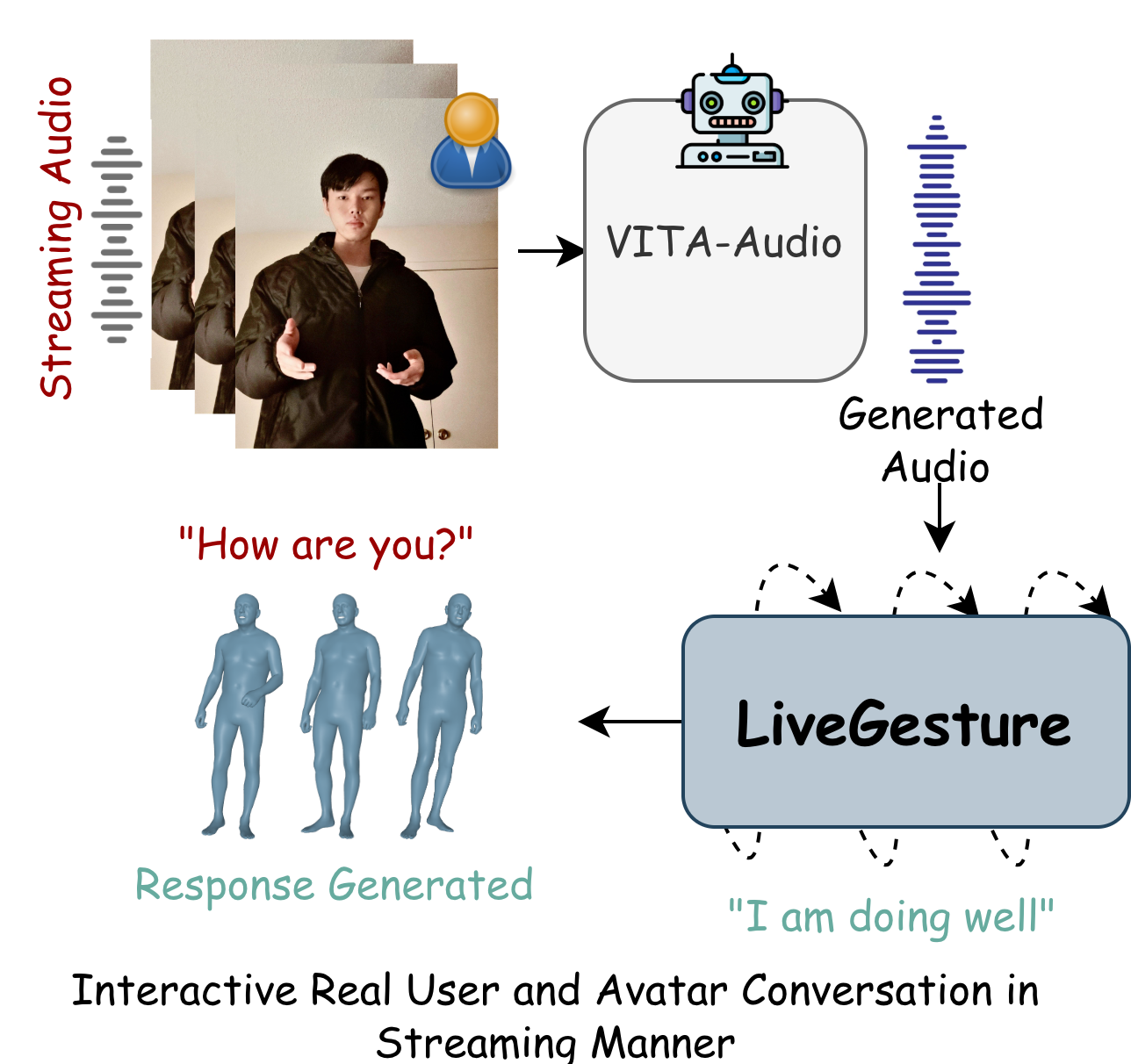}

\caption{
Interactive human–avatar conversation enabled by \emph{LiveGesture}. User speech is converted into a spoken reply by VITA-Audio, while our LiveGesture streaming gesture model simultaneously generates synchronized full-body SMPL-X motions from live audio, allowing the avatar to respond in real time.
}

    \label{fig:application}
\end{figure}

\begin{table*}[t]
\caption{Ablation studies of LiveGesture design choices, including core components, architecture, tokenization, audio encoders, loss weights, and UGR.}

  \label{tab:ab}
  \vspace{-0.1cm}
  \renewcommand{\tabcolsep}{1.8pt}
  \small

  %----------------------------------------------------
  % (a) Effectiveness of Proposed Components
  %----------------------------------------------------
  \begin{subtable}[!t]{0.3\linewidth}
  \centering
  \scalebox{0.92}{
  \begin{tabular}{lccc}
    \toprule
    {\it Configuration}  & FGD$\downarrow$  & BC$\rightarrow$ & Div.$\uparrow$  \\
    \midrule
    w/o PILOR         & 4.89  & 0.774 & 13.41 \\
    w/o UGR           & 4.98  & 0.723 & 13.64 \\
    w/o spatial attn. & 6.64  & 0.732 & 11.56 \\
    w/o temporal attn.& 15.52 & 0.712 & 10.40 \\
    w/o text cues     & 4.60  & \textbf{0.796} & 13.96 \\
    \rowcolor{gray!15} \textbf{LiveGesture (ours)} & \textbf{4.57} & 0.794 & \textbf{13.97} \\
    \bottomrule
  \end{tabular}}
  \caption{\small{Effectiveness of LiveGesture components}}
  \label{tab:ab-module}
\end{subtable}
  \hspace{\fill}
  %----------------------------------------------------
  % (b) Local Experts Performance
  %----------------------------------------------------
  \begin{subtable}[!t]{0.34\linewidth}
    \centering
    \scalebox{0.92}{
    \begin{tabular}{lccc}
      \toprule
      {\it Architecture}  & FGD$\downarrow$  & BC$\rightarrow$ & Div.$\uparrow$  \\
      \midrule
      Local region eXperts only   & 6.458 & 0.762 & 12.844 \\
      \rowcolor{gray!15} \textbf{LiveGesture (ours)} & \textbf{4.57} & \textbf{0.794} & \textbf{13.91} \\
      \bottomrule
    \end{tabular}}
    \caption{\small{Local experts vs.\ full LiveGesture}}
    \label{tab:ab-architecture}
  \end{subtable}
  \hspace{\fill}
  %----------------------------------------------------
  % (c) Streamable Motion Tokenizer Design
  %----------------------------------------------------
\begin{subtable}[!t]{0.32\linewidth}
  \centering
  \scalebox{0.92}{
  \begin{tabular}{lc}
    \toprule
    {\it Stage-2 trainable components} & FGD$\downarrow$ \\
    \midrule 
    1-stage (Enc+Dec+Quant)           & 4.691 \\
    AE + 2-stage (Dec+Quant)          & 4.633 \\
    AE + 2-stage (Dec+Quant+MLP)      & 4.610 \\
    AE + 2-stage (Quant+MLP)          & \textbf{4.557} \\
    \bottomrule
  \end{tabular}}
  \caption{\small{
    Effect of which SVQ components are trainable in Stage~2
    after autoencoder pretraining on xAR-Fuse performance.
  }}
  \label{tab:ab-time}
\end{subtable}

  \vspace{0.2cm}

  %----------------------------------------------------
  % (d) Streamable Audio Encoders
  %----------------------------------------------------
  \begin{subtable}[!t]{0.3\linewidth}
    \centering
    \scalebox{0.92}{
    \begin{tabular}{lccc}
      \toprule
      {\it Model Type} & FGD$\downarrow$ & BC$\rightarrow$ & Params (M)$\downarrow$ \\
      \midrule
      Magic-Codec~\cite{song2025magicodec}  & \textbf{4.51} & 0.793 & 103 \\
      FocalCodec~\cite{della2025focalcodec} & 4.95          & 0.773 & 88 \\
      \rowcolor{gray!15} \textbf{ours} & 4.57 & \textbf{0.794} & \textbf{0.5} \\
      \bottomrule
    \end{tabular}}
    \caption{\small{Streamable audio encoders}}
    \label{tab:ab-type}
  \end{subtable}
  \hspace{\fill}
  %----------------------------------------------------
  % (e) Loss Weights
  %----------------------------------------------------
  \begin{subtable}[!t]{0.33\linewidth}
    \centering
    \scalebox{0.92}{
    \begin{tabular}{cccc}
      \toprule
      $\lambda_{\text{local}}$ & $\lambda_{\text{fuse}}$ & FGD$\downarrow$ & BC$\rightarrow$ \\
      \midrule
      0.0 & 1.0 & 4.89 & \textbf{0.795} \\
      0.3 & 1.0 & \textbf{4.57} & 0.794 \\
      0.5 & 1.0 & 4.67 & 0.781 \\
      0.7 & 1.0 & 4.74 & 0.783 \\
      1.0 & 1.0 & 4.85 & 0.774 \\
      \bottomrule
    \end{tabular}}
    \caption{\small{Role of Local Expert Loss in $\mathcal{L}_{\text{AR}}$.}}
    \label{tab:ab-speed}
  \end{subtable}
  \hspace{\fill}
  %----------------------------------------------------
  % (f) Masking Ratio in UGR
  %----------------------------------------------------
\begin{subtable}[!t]{0.33\linewidth}
  \centering
  \scalebox{0.92}{
  \begin{tabular}{lccc}
    \toprule
    {\it UGM masking schedule}  & FGD$\downarrow$  & BC$\rightarrow$  & Div.$\uparrow$ \\
    \midrule
    $\gamma(\tau \in \mathcal{U}(0, 1))$   & 4.63 & 0.775 & 12.95 \\
    $\gamma(\tau \in \mathcal{U}(0, 0.3))$ & 4.84 & 0.792 & 13.24 \\
    $\gamma(\tau \in \mathcal{U}(0, 0.5))$ & \textbf{4.57} & \textbf{0.794} & \textbf{13.91} \\
    $\gamma(\tau \in \mathcal{U}(0, 0.7))$ & 4.67 & 0.773 & 13.60 \\
    \bottomrule
  \end{tabular}}
  \caption{\small{Effect of masking schedule in uncertainty-guided masking (UGM).}}
  \label{tab:ab-sample}
\end{subtable}

  \vspace{-5mm}
\end{table*}

\subsection{Ablation Studies}

We ablate the key components of LiveGesture to understand which design choices allow a strictly causal, zero–look-ahead model to surpass offline methods. We vary (i) fusion architecture and region conditioning, (ii) motion tokenization and audio encoding, and (iii) streaming-aware training objectives and masking strategies. Quantitative results are summarized in Table~\ref{tab:ab}. More 

\partitle{Effectiveness of LiveGesture Components}
Table~\ref{tab:ab-module} shows that fusion-level temporal attention is essential: removing it causes the largest drop across all metrics (FGD 4.57→15.52, BC 0.794→0.712, Div.\ 13.97→10.40), confirming the need for causal temporal modeling beyond local eXperts. Spatial attention also clearly improves FGD and diversity (6.64 / 11.56 without it), highlighting the importance of inter-region coordination for full-body coherence. PILOR removal degrades FGD and BC (4.89 / 0.774), indicating that low-rank pre-infusion alignment stabilizes fusion over frozen eXperts. Disabling UGR yields the worst BC (0.723) and higher FGD (4.98), showing that uncertainty-guided refinement mitigates exposure bias under streaming errors. Text removal leaves FGD and diversity almost unchanged and slightly improves BC (0.796), suggesting that audio is the primary driver of rhythm, with text providing modest semantic/style refinements.

\partitle{Region-eXperts vs.\ Full xAR-Fuse}
Table~\ref{tab:ab-architecture} compares local region-eXperts alone to the full architecture. eXperts alone achieve reasonable quality (FGD 6.458, BC 0.762, Div.\ 12.844) but underperform LiveGesture. Adding xAR-Fuse on top of frozen eXperts improves all metrics (FGD 4.57, BC 0.794, Div.\ 13.91), validating our hierarchical design: specialized region-eXperts provide strong local priors, while a dedicated causal fusion stage is needed to turn them into globally coherent streaming motion.

\partitle{Effect of SVQ on xAR-Fuse}
Table~\ref{tab:ab-time} examines which SVQ components are trainable in Stage~2. Jointly training encoder, decoder, and quantizer (1-stage) yields the worst FGD (4.691), indicating that learning compression and causal decoding simultaneously is suboptimal. In the two-stage setting, performance improves as we freeze more of the autoencoder, with the best FGD when only the Quantizer+MLP are updated and the decoder is fully frozen (4.557). This supports our SVQ design (Sec.~\ref{sec:svq}): first learn a stable, streamable latent space, then adapt a lightweight quantizer and projection on top without altering the decoder.

\partitle{Streamable Audio Encoder}
Table~\ref{tab:ab-type} compares our light-weight Steamable Audio Encoder with heavier pre-trained neural codecs. Magic-Codec slightly improves FGD (4.51 vs.\ 4.57) but uses 103M params, whereas our audio encoder attains near-identical FGD, the best BC (0.794), and only 0.5M parameters; FocalCodec is both larger and less accurate. This shows that a lightweight, task-specific, strictly causal encoder aligned to the motion token rate is sufficient to generalize well and deliver high-quality, low-latency gesture streaming without the overhead of generic codecs.

\partitle{Role of Local Expert Loss}
Table~\ref{tab:ab-speed} ablates the weights in \(\mathcal{L}_{\text{AR}} = \lambda_{\text{local}} \mathcal{L}_{\text{local}} + \lambda_{\text{fuse}} \mathcal{L}_{\text{fuse}}\). Using only fusion loss (\(\lambda_{\text{local}} = 0\)) yields good BC but worse FGD, so xAR-Fuse alone under-regularizes regional dynamics. A small local weight (\(\lambda_{\text{local}} = 0.3\)) gives the best FGD while preserving BC, whereas larger values hurt both metrics, indicating that region-eXperts work best as lightly guided priors with xAR-Fuse handling final coordination.

\partitle{Masking Ratio in Uncertainty-Guided Refinement (UGR)}
Table~\ref{tab:ab-sample} evaluates different UGR masking schedules. Very broad or very narrow masking ranges lead to worse FGD or BC, while our default schedule \(\gamma(\tau \in \mathcal{U}(0, 0.5))\) achieves the best trade-off (FGD 4.57, BC 0.794, Div.\ 13.91). This suggests that UGR is most effective when it injects moderate corruption that mimics streaming errors without making refinement unstable.

\vspace{-5pt} 
\subsection{Application: Interactive Human$-$Avatar Conversation}
LiveGesture can be plugged into streaming speech agents such as VITA-Audio~\cite{long2025vita} to enable fully interactive human–avatar conversations. In our prototype, user speech is converted by VITA-Audio into a spoken response whose waveform is streamed into our xAR-based gesture model, driving synchronized full-body SMPL-X motion in real time. The same pipeline can be extended with virtual clothing or character skins over the generated motion, enabling expressive avatars for games, VTubers, and telepresence.

\section{Conclusion}
We introduce \emph{LiveGesture}, a zero–look-ahead, fully streamable co-speech full-body gesture framework that integrates a streamable asymmetric motion tokenizer with hierarchical region-eXpert autoregression. A per-region SVQ tokenizer first produces causal motion tokens; local audio-conditioned eXperts model region-specific dynamics, while xAR-Fuse performs causal spatio-temporal fusion across regions. Hybrid uncertainty-guided masking, along with random region masking strategies, trains the model to refine low-confidence predictions under realistic streaming noise. Together, these components enable diverse, coherent, and beat-synchronous gesture generation over arbitrarily long sequences, making \emph{LiveGesture} well suited for interactive applications such as live avatars, VTubers, telepresence agents, social robots, and AR/VR characters

{
    \small
    \bibliographystyle{ieeenat_fullname}
    \bibliography{main}
}

\clearpage
\onecolumn

\setcounter{table}{0}  % Reset table numbering
\setcounter{figure}{0} % Reset figure numbering
\setcounter{section}{0} % Reset figure numbering

\section{Supplementary Material}
\appendix

\section{Overview}
The supplementary material is organized as follows:
\begin{itemize}
    \item Section~\ref{sec:implementation}: Implementation Details of the Streamable Vector-Quantized Motion Tokenizer (SVQ) and the Hierarchical Autoregressive Transformer (HAR)
    \item Section~\ref{sec:metrics}: Evaluation Metrics
    \item Section~\ref{sec:sota_face}: SOTA Comparison: Quantitative Results Without Face Module
    \item Section~\ref{sec:sota_user_stdy}: SOTA Comparison: User Study
    \item Section~\ref{sec:codebook_size}: Impact of Codebook Size in Streamable Vector-Quantized Motion Tokenizer
    \item Section~\ref{sec:ca_sa_causal}: Causal Attention Design in xAR and xAR-Fuse

    \item Section~\ref{sec:cfg_role}: Role of Classifier-Free Guidance in \emph{LiveGesture}
    \item Section~\ref{sec:svq_compositional}: Role of Compositional vs.\ Full-Body SVQ Tokenization
    \item Section~\ref{sec:region_mask}: Ablation on Role of Region Masking in Fusion Training
    \item Section~\ref{sec:region_noise}: Impact of Noise Injection in Local Region-eXperts
    \item Section~\ref{sec:ablation_attention_depth}: Ablation on Attention Depth in xAR and xAR-Fuse
\end{itemize}

\section{Implementation Details}
\label{sec:implementation}

\emph{LiveGesture} is implemented in PyTorch and trained in two major phases: (i) the per-region \emph{Streamable Vector-Quantized Motion Tokenizer} (SVQ), and (ii) the \emph{Hierarchical Autoregressive Transformer} (HAR).

\subsection{Streamable Vector-Quantized Motion Tokenizer}
\label{sec:impl_svq}

\partitle{Asymmetric Autoencoder Pretraining.}
We train one \emph{Streamable Autoencoder} (SA) per SMPL-X body region, $\mathcal{R}=\{\text{upper body}, \text{lower body}, \text{hands}, \text{face}\}$. The input parameterization for each region is as follows: 13 upper-body joints (78-D Rot6D), 30 hand joints (180-D Rot6D), 100-D FLAME expression parameters plus 3-D jaw rotation for the face, and 9 lower-body joints (54-D Rot6D) augmented with global translation (3-D) and four binary foot-contact indicators. Global translation and contact flags are included only in the lower-body region, as leg motion and foot contacts provide strong supervision for root displacement and help reduce foot-sliding artifacts.

For training, each SA processes motion in fixed sliding windows of length $T_w = 16$ frames, denoted $\{\boldsymbol{\theta}^{\text{region}}_t\}_{t=1}^{T_w}$ for a given region. The bidirectional encoder $E$ begins with a temporal convolutional stem that lifts framewise SMPL-X inputs into a higher-dimensional feature space. It then applies two downsampling stages, each consisting of a stride-2 temporal convolution (halving the frame rate) followed by a \texttt{ResNet1D} block composed of dilated temporal residual layers. These residual layers jointly capture local kinematic patterns and longer-range bidirectional dependencies within the window. After two stages, the encoder outputs a latent sequence
\[
z^{\text{region}} = \{z_\tau\}_{\tau=1}^{T_w/4},
\]
i.e., a compact motion representation at one-quarter of the original frame rate. This asymmetric design allows $E$ to exploit both past and future frames to build a globally coherent latent space, while the decoder remains strictly causal.

The Causal Stream Decoder $D_{\text{CS}}$ mirrors the encoder hierarchy but enforces strict causality: all convolutions use left-only padding, so the reconstruction at frame $t$ depends only on latent tokens $\{z^{\text{region}}_{1:\tau}\}$ up to the corresponding downsampled index. Each upsampling stage performs nearest-neighbor temporal upsampling by~2, followed by a causal \texttt{ResNet1D} block that refines the feature sequence using only past context. After two stages, the decoder reconstructs the full-resolution regional motion window
\[
\hat{\boldsymbol{\theta}}^{\text{region}}
=
D_{\text{CS}}(z^{\text{region}}).
\]
Because $D_{\text{CS}}$ is strictly causal and operates at a fixed downsampling factor, the same architecture can be run convolutionally over arbitrarily long sequences at inference time without violating the zero--look-ahead constraint. In the final streaming system, only the causal decoder $D_{\text{CS}}$ is used at inference; the bidirectional encoder $E$ is employed solely during offline training to shape the latent space.

Each SA is trained independently with per-region normalization and an L1 reconstruction loss
\[
\mathcal{L}_{\text{AE}}
=
\lambda_{\text{AE}}
\left\|
\hat{\boldsymbol{\theta}}^{\text{region}}
-
\boldsymbol{\theta}^{\text{region}}
\right\|_1,
\quad
\lambda_{\text{AE}} = 1.
\]
This stage focuses solely on learning a temporally coherent, streamable latent space and does not involve any vector quantization.

\partitle{Quantization Learning.}
In Stage~2, we convert the continuous latents $z^{\text{region}}$ into discrete, time-synchronous SVQ motion tokens while preserving the temporal geometry and causal decoding behavior learned in Stage~1. To this end, we freeze both the bidirectional encoder $E$ and the causal decoder $D_{\text{CS}}$ and learn a region-specific vector-quantized tokenizer on top of the SA latents.

For each region $r \in \mathcal{R}$, we introduce a codebook
\[
C^{\text{region}} = \{c_k\}_{k=1}^{K}
\in \mathbb{R}^{K \times 128},
\]
with $K=2048$ entries, updated using EMA with decay $0.99$. The encoder latents $z^{\text{region}}_\tau$ are first linearly projected into the 128-D code space. Each projected latent is then assigned to its nearest codebook vector $c_k$, producing a discrete token index and the corresponding dequantized embedding sequence
\[
\hat{z}^{\text{region}} = \{\hat{z}_\tau\}_{\tau=1}^{T_w/4}.
\]

To remain compatible with the frozen decoder latent space, each region employs a lightweight projection head $W^{\text{region}}$, implemented as a small MLP. This head maps dequantized latents back into the latent space expected by $D_{\text{CS}}$:
\[
\tilde{z}^{\text{region}}
=
\{ \tilde{z}_\tau \}_{\tau=1}^{T_w/4},
\quad
\tilde{z}_\tau = W^{\text{region}}(\hat{z}_\tau).
\]
The projected sequence is then decoded causally to reconstruct the regional motion window,
\[
\hat{\boldsymbol{\theta}}^{\text{region}}
=
D_{\text{CS}}(\tilde{z}^{\text{region}}).
\]

In this stage, only the codebook $C^{\text{region}}$ and projection head $W^{\text{region}}$ are trainable; $E$ and $D_{\text{CS}}$ remain fixed. The codebook is maintained with EMA and we periodically reset rarely used entries to avoid codebook collapse and encourage effective usage of the discrete space. The Stage~2 objective combines an L1 reconstruction term on SMPL-X region parameters with a standard VQ codebook loss:
\[
\mathcal{L}_{\text{stage2}}
=
\lambda_{\text{rec}}
\left\|
\hat{\boldsymbol{\theta}}^{\text{region}}
-
\boldsymbol{\theta}^{\text{region}}
\right\|_1
+
\lambda_{\text{cb}}\,
\mathcal{L}_{\text{cb}}(z^{\text{region}}, e^{\text{region}}),
\]
where $e^{\text{region}}$ denotes the selected codebook embeddings, $\lambda_{\text{rec}}=1$, and $\lambda_{\text{cb}}=0.2$. Gradients pass through the quantizer via a straight-through estimator. This two-stage asymmetric design ensures that (i) the latent space and causal decoder remain stable and well-conditioned for streaming, and (ii) the SVQ tokens are compact, region-specific, and time-synchronous at one-quarter of the original motion frame rate, making them ideal discrete inputs for the downstream region-eXpert autoregressive transformers in HAR.

\subsection{Hierarchical Autoregressive Transformer (HAR)}
\label{sec:impl_har}

\partitle{Region-eXpert Autoregressive Transformer (xAR).}
Each \emph{region-eXpert} xAR module performs causal autoregressive modeling on the SVQ motion tokens produced at one-quarter of the original motion frame rate. For every region $r \in \mathcal{R}=\{\text{upper body},\text{lower body},\text{hands},\text{face}\}$, we maintain an independent stream of SVQ tokens together with aligned audio tokens from the streamable audio encoder and optional text tokens. The model operates with a maximum history of $32$ past motion tokens, each mapped to a 128-D codebook embedding ($K=2048$ entries) and projected to a 256-D representation through a small MLP; rotary positional embeddings are added for stable temporal alignment during streaming. Each region-eXpert is implemented as a lightweight causal Transformer with $L_{\text{xAR}}=2$ blocks, where every block contains three \emph{causal audio--motion cross-attention} layers that attend only to past and current audio/text tokens, followed by three \emph{causal temporal self-attention} layers applied over the region’s token history; all attention layers use strict lower-triangular masks to ensure zero--look-ahead. The final hidden state at time $t$ is passed through a region-specific linear classifier to produce logits over the 2048-entry codebook vocabulary. To improve robustness, we inject Gaussian noise with standard deviation $\sigma_{\text{noise}}=0.1$ into the embedded motion history with probability $p_{\text{noise}}=0.2$, and apply classifier-free dropout to audio/text inputs with probability $p_{\text{cf}}=0.1$, enabling classifier-free guidance at inference with scale $\gamma=1.25$. All four region-eXperts share the same streamable audio encoder but maintain independent Transformer and classifier weights, and are trained using Adam (learning rate $1\times 10^{-4}$, batch size 128) under the standard autoregressive objective.

\vspace{6pt}
\partitle{Causal Spatial--Temporal Fusion (xAR-Fuse).}

xAR-Fuse enforces whole-body coordination by operating on top of the frozen hidden states produced by the region-eXpert xAR modules. At each time step $t$, we gather the region-wise features $\{h^{r}_t\}_{r\in\mathcal{R}}$ and align them using lightweight per-region PILOR adapters, each implemented as a single linear layer with a residual connection. These adapters add only a small number of parameters per region while reliably mapping independently learned region features into a shared fusion space. The aligned features are then processed by a causal fusion Transformer with $L_{\text{fuse}}=3$ blocks. Each fusion block contains three components: (i) \emph{inter-region spatial attention} across regions at the current time step, (ii) \emph{causal global temporal attention} with key--value caching for long-horizon streaming, and (iii) \emph{causal audio--motion cross-attention} that conditions fused region tokens on past and current audio/text input. The resulting fused representations are fed into the same region-specific token classifiers used in the local xAR stage, now predicting SVQ tokens from joint multi-region context.

Training uses the hybrid masking strategy described in the main method: uncertainty-guided token masking (UGM) corrupts a subset of the lowest-confidence tokens according to a cosine schedule $\lambda_{\text{UGR}}(s)$, where the effective masking ratio increases from $0$ to a maximum of $0.5$ over the course of training; for each batch, a masking ratio is sampled uniformly in $[0,\lambda_{\text{UGR}}(s)]$ and applied to the selected tokens. Random region masking (RM) is implemented in an analogous way: we gradually increase a region-drop probability $p_{\text{drop}}$ from $0$ to $0.5$, and for each batch sample a value of $p_{\text{drop}}$ within the current range and use it as the probability of dropping an entire region’s motion-token sequence, encouraging cross-region reasoning under missing modalities. We also apply classifier-free dropout with probability $p_{\text{cf}}=0.1$, identical to xAR, to retain compatibility with inference-time classifier-free guidance. During xAR-Fuse training, the SVQ tokenizer, causal audio encoder, and all xAR eXperts are frozen; only the PILOR adapters, fusion Transformer blocks, and token classifiers are updated. We train xAR-Fuse using Adam (learning rate $1\times 10^{-4}$, batch size 128). Because all attention operations are strictly causal, the resulting model runs directly in real-time streaming mode without any architectural changes. In the overall training objective, we weight the local xAR loss and the fusion loss with coefficients $\lambda_{\text{local}} = 0.3$ and $\lambda_{\text{fuse}} = 1.0$, respectively.

\subsection{Application: Interactive Human--Avatar Conversation}
\label{sec:impl_application}

For deployment, \emph{LiveGesture} connects directly to a streaming speech system such as VITA-Audio. Audio is emitted in small chunks (200\,ms hop) and immediately passed to the causal audio encoder, which updates the audio tokens at the SVQ rate. HAR advances synchronously, generating one SVQ token per region at each audio step. The SVQ decoder reconstructs full SMPL-X poses in real time, enabling fully synchronized speech--gesture behavior. Because the entire HAR stack is strictly causal, the same implementation supports real-time human--avatar interaction without look-ahead or buffering.

\section{Evaluation Metrics}
\label{sec:metrics}

We evaluate \emph{LiveGesture} using four standard metrics that measure realism, variability, speech--motion synchrony, and facial accuracy, following the official BEAT2 protocol.

\partitle{Fr\'echet Gesture Distance (FGD).}
FGD~\cite{yoon2020speech} measures the distributional similarity between real and generated full-body motion in the feature space of a pretrained gesture encoder. Let feature sets extracted from real and generated gestures be
\(
G = \{\mathbf{g}_i\}
\)
and
\(
\hat{G} = \{\hat{\mathbf{g}}_i\}
\),
with means
\(
\boldsymbol{\mu}_{G},\ \boldsymbol{\mu}_{\hat{G}}
\)
and covariances
\(
\Sigma_{G},\ \Sigma_{\hat{G}}.
\)
FGD is defined as
\begin{equation}
\mathrm{FGD}(G,\hat{G}) =
\|\boldsymbol{\mu}_{G} - \boldsymbol{\mu}_{\hat{G}}\|_2^{2}
+
\mathrm{Tr}\!\left(
\Sigma_{G} + \Sigma_{\hat{G}}
- 2(\Sigma_{G}\Sigma_{\hat{G}})^{1/2}
\right).
\end{equation}
Lower FGD indicates that generated gestures follow natural human-motion statistics, which is critical for strictly causal, zero--look-ahead generation.

\partitle{L1 Diversity.}
L1 Diversity~\cite{li2021audio2gestures} measures variability across multiple gesture realizations produced for the same audio. Given \(N\) generated sequences with corresponding joint positions
\(
\{\mathbf{p}^{(i)}_t\}_{t=1}^{T}
\)
for $i=1,\dots,N$, the diversity score is
\begin{equation}
\mathrm{Div.} =
\frac{1}{2N(N-1)T}
\sum_{i=1}^{N}
\sum_{\substack{j=1 \\ j \neq i}}^{N}
\sum_{t=1}^{T}
\bigl\|
\mathbf{p}^{(i)}_t - \mathbf{p}^{(j)}_t
\bigr\|_{1}.
\end{equation}
Global translation of the SMPL-X body is removed prior to evaluation. Higher values indicate more expressive and varied motion under causal token-by-token prediction.

\partitle{Beat Constancy (BC).}
Beat Constancy~\cite{li2021ai} evaluates the synchrony between motion beats and prosodic beats in the audio. Motion beats are detected from local minima of upper-body joint velocity, while audio beats correspond to peaks in prosodic intensity. Let
\(g_{\mathrm{mot}}\)
and
\(a_{\mathrm{aud}}\)
denote the sets of detected motion and audio beat times. BC is computed as
\begin{equation}
\mathrm{BC} =
\frac{1}{|g_{\mathrm{mot}}|}
\sum_{b_g \in g_{\mathrm{mot}}}
\exp\!\left(
-\frac{
\min_{b_a \in a_{\mathrm{aud}}}
\|b_g - b_a\|^{2}
}{2\sigma^{2}}
\right),
\end{equation}
where $\sigma$ is a temporal tolerance parameter. Higher BC indicates stronger speech--gesture alignment. Because \emph{LiveGesture} is fully causal with zero--look-ahead, BC directly reflects its ability to track prosody in real time.

\partitle{Facial Vertex MSE.}
For SMPL-X facial motion accuracy, we compute the mean squared error between ground-truth and predicted mesh vertices following~\cite{xing2023codetalker}. Let
\(
\mathbf{V} = \{\mathbf{v}_i\}
\)
and
\(
\hat{\mathbf{V}} = \{\hat{\mathbf{v}}_i\}
\)
be the sets of ground-truth and predicted facial vertices. The metric is
\begin{equation}
\mathrm{MSE}_{\mathrm{face}} =
\frac{1}{|\mathbf{V}|}
\sum_{i=1}^{|\mathbf{V}|}
\|\mathbf{v}_i - \hat{\mathbf{v}}_i\|_2^{2}.
\end{equation}
This complements body-motion metrics by evaluating fine-grained facial deformation fidelity.

\section{SOTA Comparison: Quantitative Results Without Face Module}
\label{sec:sota_face}

\begin{table*}[ht]
    \caption{Comparison with state-of-the-art methods on BEAT2 without the facial motion module. 
    \emph{LiveGesture} remains the only zero--look-ahead streaming model while achieving competitive or superior performance in BC and Diversity.}
    \label{tab:beat_no_face}
    \centering
    \scalebox{1}{
    \begin{tabular}{l|lcccc}
    \toprule
    Methods & Venue & Streaming & FGD ($\downarrow$) & BC ($\rightarrow$) & Diversity ($\uparrow$) \\
    \midrule
    \rowcolor{gray!10}
    \multicolumn{6}{c}{Offline Solutions} \\
    \midrule
    GestureLSM~\cite{liu2025gesturelsm}
        & ICCV'25
        & \xmark
        & \textbf{4.088}
        & 0.714
        & \underline{13.24} \\
    \midrule
    \rowcolor{gray!10}
    \multicolumn{6}{c}{Streaming Solution (Ours)} \\
    \midrule
    \rowcolor{green!8} \emph{LiveGesture} (ours)
        & Ours
        & \cmark
        & \underline{4.51}
        & \textbf{0.783}
        & \textbf{13.31} \\
    \bottomrule
    \end{tabular}}
    \vspace{-0.3cm}
\end{table*}

Table~\ref{tab:beat_no_face} reports quantitative results when evaluating only full-body motion without the facial module. GestureLSM, an offline model with full future context, achieves the lowest FGD by leveraging non-causal temporal information. In contrast, \emph{LiveGesture} operates under strict zero--look-ahead constraints, yet obtains the best BC and highest Diversity, indicating superior rhythm alignment and richer motion variability. This behavior reflects the strength of \emph{LiveGesture}'s causal audio conditioning and hierarchical xAR--xAR-Fuse architecture, which jointly enable expressive, beat-synchronous gestures even without access to future frames. Although the absence of facial cues slightly increases the distributional distance (FGD) for our model, the strong improvements in synchrony and diversity demonstrate that \emph{LiveGesture} maintains high perceptual quality while remaining the only fully streamable solution.

\section{SOTA Comparison: User Study}
\label{sec:sota_user_stdy}

\partitle{Protocol}
We recruit fifteen participants and present them with a mixed set of gesture clips generated by CaMN~\cite{liu2022beat}, EMAGE~\cite{liu2023emage}, GestureLSM~\cite{liu2025gesturelsm}, and \emph{LiveGesture}. Each clip depicts a speaking subject with full-body SMPL-X motion driven by BEAT2 test utterances. All videos are anonymized and randomly ordered so that participants cannot identify which method produced which sequence, reducing bias toward any specific model.

For every clip, participants provide ratings on a five-point Likert scale (1 = lowest, 5 = highest) along three criteria. \emph{Realness} evaluates the naturalness and plausibility of the produced motion. \emph{Speech--gesture synchrony} measures how well the gestures align with the rhythm and prosodic structure of the audio. \emph{Smoothness} assesses temporal continuity and penalizes jitter, discontinuities, or incoherent regional coordination. We report Mean Opinion Scores (MOS) averaged across all participants and clips for each method.

\partitle{Results}
Table~\ref{tab:user_study_sota} summarizes the MOS results. Both \emph{LiveGesture} and GestureLSM are clearly preferred over CaMN and EMAGE across all criteria, indicating that recent models produce noticeably more convincing co-speech motion. GestureLSM, an offline method with full-sequence access, achieves slightly higher Realness (4.2 vs.\ 4.1) and Smoothness (4.3 vs.\ 3.9) than \emph{LiveGesture}, reflecting its advantage in long-range temporal refinement when future frames are available. In contrast, \emph{LiveGesture} attains the best speech--gesture synchrony score (4.3), surpassing all offline baselines and aligning with its superior BC metric in the quantitative evaluation. This pattern suggests that our strictly causal xAR--xAR-Fuse architecture, together with the streamable audio encoder, is particularly effective at tracking prosody and timing under zero--look-ahead constraints, while still delivering realism and smoothness comparable to the strongest offline system. Overall, the user study confirms that a fully streaming model can match or nearly match the perceptual quality of state-of-the-art offline methods, while providing better perceived synchrony with speech.

\begin{table}[t]
    \caption{Mean Opinion Scores (MOS, 1--5, higher is better) from the user study on BEAT2 test clips. 
    The \emph{Streaming} column indicates whether the method supports zero--look-ahead streaming (\cmark) or is offline-only (\xmark). 
    \emph{LiveGesture} is the only strictly streaming model and is preferred on speech--gesture synchrony while remaining competitive in realness and smoothness.}
    \label{tab:user_study_sota}
    \centering
    \scalebox{0.95}{
    \begin{tabular}{l|cccc}
    \toprule
    Method & Streaming & Realness $\uparrow$ & Synchrony $\uparrow$ & Smoothness $\uparrow$ \\
    \midrule
    CaMN~\cite{liu2022beat}              & \xmark & 3.3 & 3.2 & 3.6 \\
    EMAGE~\cite{liu2023emage}            & \xmark & 3.5 & 3.5 & 3.7 \\
    GestureLSM~\cite{liu2025gesturelsm}  & \xmark & \textbf{4.2} & 4.1 & \textbf{4.3} \\
    \rowcolor{gray!10} \emph{LiveGesture} (ours) & \cmark & 4.1 & \textbf{4.3} & 3.9 \\
    \bottomrule
    \end{tabular}}
    \vspace{-0.3cm}
\end{table}

\section{Impact of Codebook Size in Streamable Vector-Quantized Motion Tokenizer}
\label{sec:codebook_size}

\begin{table}[ht]
    \caption{Effect of codebook size in the SVQ motion tokenizer on full-body gesture generation. 
    Larger codebooks increase representational capacity and yield consistent gains across all metrics.}
    \label{tab:codebook_size}
    \centering
    \scalebox{1}{
    \begin{tabular}{l|ccc}
    \toprule
    Codebook size & FGD ($\downarrow$) & BC ($\rightarrow$) & Diversity ($\uparrow$) \\
    \midrule
    512 entries   & 6.63 & 0.734 & 12.14 \\
    1024 entries  & 5.13 & 0.774 & 13.21 \\
    2048 entries  & 4.51 & 0.794 & 13.91 \\
    \bottomrule
    \end{tabular}}
    \vspace{-0.3cm}
\end{table}

Table~\ref{tab:codebook_size} examines how the size of the region-specific codebook $C^{\text{region}}=\{c_k\}_{k=1}^{K}$ in the SVQ motion tokenizer affects downstream gesture generation. With a small codebook ($K{=}512$), many continuous latents $z^{\text{region}}_\tau$ are mapped to the same discrete code, leading to quantization collisions that restrict the expressiveness of the motion token sequence $\{x^{r}_t\}$. This loss of granularity degrades realism (higher FGD) and weakens prosodic alignment (lower BC), as subtle variations in timing and amplitude cannot be preserved. Increasing the codebook to $K{=}1024$ reduces collisions and provides a richer set of motion primitives, improving both synchrony and diversity. The best performance is obtained with $K{=}2048$, where the token inventory is sufficiently large to capture finer-grained spatial and temporal variations while still remaining learnable under causal decoding. The resulting discrete representation enables xAR and xAR-Fuse to model expressive region-level dynamics more faithfully, yielding improved realism (FGD), stronger rhythm alignment (BC), and greater motion variability (Diversity).

\begin{table}[ht]
    \caption{
    Comparison of causal self-attention and causal audio--motion cross-attention in the Hierarchical Autoregressive Transformer (HAR). 
    All models maintain strictly aligned tokenization rates between audio tokens $\{a_t\}$ and motion tokens $\{x^{r}_t\}$ for fair comparison.
    }
    \label{tab:causal_attn}
    \centering
    \scalebox{1}{
    \begin{tabular}{l|ccc}
    \toprule
    Method                 & FGD ($\downarrow$) & BC ($\rightarrow$) & Diversity ($\uparrow$) \\
    \midrule
    Causal self-attention  & 4.63               & 0.784              & 12.53 \\
    Causal cross-attention & \textbf{4.57}      & \textbf{0.794}     & \textbf{13.91} \\
    \bottomrule
    \end{tabular}}
    \vspace{-0.3cm}
\end{table}

\section{Causal Attention Design in xAR and xAR-Fuse}
\label{sec:ca_sa_causal}

Table~\ref{tab:causal_attn} compares two causal attention strategies used in the hierarchical autoregressive transformer. 
In the ``causal self-attention'' variant, audio and motion embeddings at each time step are combined by elementwise addition,
\[
u_t = x^{r}_t + a_t,
\]
and the transformer attends only over the fused sequence $u_{1:t}$. 
While this preserves causality and keeps audio and motion aligned at the same token rate, the additive fusion collapses modality structure: the model cannot distinguish which features originate from audio and which from motion, and it cannot form directed cross-modal queries. 
As a result, the network must implicitly disentangle prosodic cues from the blended representation, which weakens rhythm conditioning and reduces expressive variability, leading to higher FGD (4.63), lower BC (0.784), and reduced Diversity (12.53). 

In contrast, the ``causal cross-attention'' variant maintains separate motion queries and audio keys/values, enabling each motion token $x^{r}_t$ to directly attend to synchronized audio cues $\{a_{1:t}\}$ through an explicit cross-modal pathway. 
This structured alignment allows the model to selectively extract energy changes, prosodic beats, and local speech dynamics essential for gesture timing and expressiveness. 
Consequently, causal cross-attention yields systematically better results (FGD = 4.57, BC = 0.794, Div.~= 13.91), showing that explicit causal audio--motion conditioning is more effective than additive fusion for real-time gesture generation.

\section{Role of Classifier-Free Guidance in LiveGesture}
\label{sec:cfg_role}

\begin{table}[ht]
    \caption{Effect of classifier-free guidance scale $\gamma$ on streaming gesture generation.}
    \label{tab:cfg_ablation}
    \centering
    \begin{tabular}{c|ccc}
    \toprule
    $\gamma$ & FGD$\downarrow$ & BC$\rightarrow$ & Div.$\uparrow$ \\
    \midrule
    1.00  & 4.61 & 0.781 & 12.90 \\
    1.25  & 4.57 & 0.794 & 13.91 \\
    1.35  & 5.23 & 0.763 & 12.80 \\
    2.00  & 6.42 & 0.756 & 11.74 \\
    \bottomrule
    \end{tabular}
    \vspace{-0.2cm}
\end{table}

Table~\ref{tab:cfg_ablation} examines how the classifier-free guidance scale $\gamma$ affects the token prediction distribution in \emph{LiveGesture}. During inference, the guided logits for each region $r$ are computed as
\[
\ell_{\text{guided}}^{r} = 
\ell_{\text{uncond}}^{r}
+ \gamma\,\bigl(\ell_{\text{cond}}^{r} - \ell_{\text{uncond}}^{r}\bigr),
\]
which amplifies the influence of audio-conditioned predictions while preserving causal decoding. When $\gamma$ is too small ($\gamma = 1.00$), the model underutilizes modality conditioning, resulting in weaker synchronization and reduced expressiveness. Moderate guidance ($\gamma = 1.25$) provides the best balance, sharpening the conditional distribution enough to strengthen prosodic alignment (highest BC) and support rich gesture variability (highest Diversity) without oversuppressing natural motion variability, leading to the lowest FGD. Increasing $\gamma$ further ($\gamma \geq 1.35$) over-amplifies conditional logits, causing overly deterministic predictions that reduce Diversity and destabilize temporal dynamics, which ultimately harms realism and synchrony under streaming constraints. These results demonstrate that \emph{LiveGesture} benefits from moderate classifier-free guidance, which reinforces audio--motion coupling while maintaining the flexibility needed for expressive full-body gestures.

\section{Role of Compositional vs.\ Full-Body SVQ Tokenization}
\label{sec:svq_compositional}

\begin{table}[ht]
    \caption{Comparison between a single full-body SVQ tokenizer and compositional per-region SVQ tokenizers. Both variants use the same total codebook capacity (2048 entries with 128-d embeddings).}
    \label{tab:svq_compositional}
    \centering
    \begin{tabular}{l|ccc}
    \toprule
    Method & FGD$\downarrow$ & BC$\rightarrow$ & Div.$\uparrow$ \\
    \midrule
    Full-body SVQ (1 tokenizer)    & 6.84 & 0.753 & 11.23 \\
    Per-region SVQ (4 tokenizers)  & 4.57 & 0.794 & 13.91 \\
    \bottomrule
    \end{tabular}
    \vspace{-0.2cm}
\end{table}

Table~\ref{tab:svq_compositional} compares two approaches for streamable motion tokenization under identical codebook capacity (\(2048 \times 128\)). A single full-body SVQ must quantize the entire SMPL-X pose vector into a single latent stream \(z_{\tau}\), forcing one codebook \(C\) to represent heterogeneous motion patterns spanning upper body, lower body, hands, and face. This produces severe quantization interference: high-frequency regions (e.g., hands) and low-frequency regions (e.g., torso) compete for the same discrete codes, leading to token collisions and loss of fine-grained structure. As a result, the downstream autoregressive models receive less informative tokens \(x_t\), which degrades realism (higher FGD), weakens prosodic synchronization (lower BC), and suppresses expressive variability (lower Diversity). 

In contrast, the compositional design factorizes the motion stream into region-specific latent sequences \(z^{r}_{\tau}\) with their own codebooks \(C^{\text{region}}\). This specialization enables each SVQ to capture the appropriate temporal and spatial scale of its region without cross-region interference. The resulting tokens $x_t^{r}$ preserve fine-grained dynamics and yield much richer conditioning signals for xAR and xAR-Fuse, improving FGD, BC, and Diversity as shown in Table~\ref{tab:svq_compositional}.

\section{Ablation on Role of Region Masking in Fusion Training}
\label{sec:region_mask}

\begin{table}[ht]
    \caption{Effect of random region masking (RM) on fusion training. 
    $p_{\text{drop}} \sim \mathcal{U}(0, a)$ denotes the range of probabilities used to fully mask a region's token trajectory.}
    \label{tab:region_mask}
    \centering
    \begin{tabular}{c|ccc}
    \toprule
    $p_{\text{drop}}$ range & FGD$\downarrow$ & BC$\rightarrow$ & Div.$\uparrow$ \\
    \midrule
    $\mathcal{U}(0, 0.2)$ & 4.57 & 0.794 & 13.91 \\
    $\mathcal{U}(0, 0.3)$ & 4.85 & 0.753 & 13.14 \\
    $\mathcal{U}(0, 0.5)$ & 5.30 & 0.770 & 12.82 \\
    \bottomrule
    \end{tabular}
    \vspace{-0.3cm}
\end{table}

Table~\ref{tab:region_mask} analyzes the effect of region-level masking in Stage~2 of fusion training, where an entire region's token history $\{x^{r}_{1:t}\}$ is removed with probability $p_{\text{drop}} \sim \mathcal{U}(0,a)$. Moderate masking ($a = 0.2$) yields the best performance because it gently exposes the fusion transformer to incomplete cross-region cues while preserving enough valid context to learn stable spatio--temporal dependencies among regions. As $a$ increases, masking removes critical information from multiple regions, forcing xAR-Fuse to infer missing motion solely from $\{a_{1:t}, w_{1:t}\}$ and the remaining regions. This degrades the quality of the fused representations $\tilde{h}^{r}_t$ and weakens whole-body coordination, leading to reduced synchrony (lower BC), reduced variability (lower Diversity), and larger distribution drift (higher FGD). These results confirm that region masking is beneficial only when corruption remains mild, allowing the fusion model to learn robustness without collapsing inter-region dynamics.

\section{Impact of Noise Injection in Local Region-eXperts}
\label{sec:region_noise}

\begin{table}[ht]
    \caption{Effect of noise injection in local region-eXperts during Stage~1 training. 
    $p_{\text{noise}} \sim \mathcal{U}(0, a)$ determines the probability of adding Gaussian noise to embedded history tokens.}
    \label{tab:local_noise}
    \centering
    \begin{tabular}{c|ccc}
    \toprule
    $p_{\text{noise}}$ range & FGD$\downarrow$ & BC$\rightarrow$ & Div.$\uparrow$ \\
    \midrule
    $\mathcal{U}(0, 0.2)$ & 4.57 & 0.794 & 13.91 \\
    $\mathcal{U}(0, 0.3)$ & 4.67 & 0.773 & 13.01 \\
    $\mathcal{U}(0, 0.5)$ & 5.30 & 0.760 & 12.63 \\
    \bottomrule
    \end{tabular}
    \vspace{-0.3cm}
\end{table}

Table~\ref{tab:local_noise} evaluates the effect of noise injection into the embedded history tokens of each region-eXpert, where noise is applied with probability $p_{\text{noise}} \sim \mathcal{U}(0,a)$. Light noise ($a = 0.2$) improves robustness by preventing overreliance on perfectly clean token histories, which rarely occur during causal autoregressive inference. This helps each local eXpert learn stable conditional distributions
\[
p_\phi^{r}\bigl(x_t^{r} \mid x_{1:t-1}^{r}, a_{1:t}, w_{1:t}\bigr),
\]
resulting in better synchrony and variance during downstream fusion. However, larger $a$ introduces excessive corruption early in training, degrading the temporal structure in the latent sequences and disrupting the mapping between region tokens and audio cues. This harms both BC and Diversity, and increases FGD due to over-regularization. These results show that mild stochastic perturbation is sufficient to improve streaming robustness, whereas heavy noise erodes the fine-grained temporal patterns essential for expressive gesture generation.

\section{Ablation on Attention Depth in xAR and xAR-Fuse}
\label{sec:ablation_attention_depth}

\begin{table}[ht]
    \caption{Effect of causal audio--motion cross-attention depth and causal temporal self-attention depth in the Region-eXpert Autoregressive Transformer (xAR).}
    \label{tab:xar_attention_depth}
    \centering
    \scalebox{1}{
    \begin{tabular}{c|c|ccc}
    \toprule
    Cross-attn layers & Temporal-attn layers & FGD$\downarrow$ & BC$\rightarrow$ & Div.$\uparrow$ \\
    \midrule
    2 & 2 & 4.60 & 0.791 & 13.42 \\
    3 & 3 & \textbf{4.57} & \textbf{0.794} & \textbf{13.91} \\
    \bottomrule
    \end{tabular}}
    \vspace{-0.2cm}
\end{table}

\begin{table}[ht]
    \caption{Effect of causal audio--motion cross-attention depth and causal spatial--temporal attention depth in the fusion transformer (xAR-Fuse).}
    \label{tab:xarfuse_attention_depth}
    \centering
    \scalebox{1}{
    \begin{tabular}{c|c|ccc}
    \toprule
    Cross-attn layers & Spatio--temporal layers & FGD$\downarrow$ & BC$\rightarrow$ & Div.$\uparrow$ \\
    \midrule
    2 & 2 & 4.65 & 0.784 & 13.35 \\
    3 & 3 & \textbf{4.57} & \textbf{0.794} & \textbf{13.91} \\
    \bottomrule
    \end{tabular}}
    \vspace{-0.2cm}
\end{table}

Tables~\ref{tab:xar_attention_depth} and~\ref{tab:xarfuse_attention_depth} show the effect of increasing the depth of causal attention in both the local Region-eXpert transformers (xAR) and the global fusion transformer (xAR-Fuse). 

In xAR, raising the number of causal audio--motion cross-attention layers and causal temporal self-attention layers from two to three consistently improves FGD, BC, and Diversity. Under strict zero--look-ahead constraints, deeper causal modeling allows each expert to more effectively capture fine-grained dependencies in the joint sequence $(x^{r}_{1:t-1}, a_{1:t})$, improving rhythm sensitivity and region-specific temporal expressiveness.

A similar trend is observed for xAR-Fuse. Increasing the number of causal cross-attention layers and spatio--temporal fusion layers from two to three enhances FGD, BC, and Diversity. The deeper configuration performs more rounds of causal spatial reasoning over region embeddings $\{h^{r}_t\}$ and more extensive temporal reasoning over global motion history, resulting in stronger whole-body coordination and improved audio--motion synchrony.

Overall, the deeper (3+3)-layer configuration yields the best performance in both xAR and xAR-Fuse. Under strictly causal conditions, both local and global modules benefit from increased attention depth, compensating for the absence of future context and producing coherent, expressive, rhythm-aligned motion in real time.

% {
%     \small
%     \bibliographystyle{ieeenat_fullname}
%     \bibliography{main}
% }

\end{document}